\documentclass[11pt,a4paper]{article}
\usepackage{times,latexsym}
\usepackage{url}
\usepackage[T1]{fontenc}

\usepackage[acceptedWithA]{tacl2018v2}
\usepackage{xspace,mfirstuc,tabulary}

\newif\iftaclinstructions
\taclinstructionsfalse % AUTHORS: do NOT set this to true
\iftaclinstructions
\renewcommand{\confidential}{}
\renewcommand{\anonsubtext}{(No author info supplied here, for consistency with
TACL-submission anonymization requirements)}
\newcommand{\instr}
\fi

\iftaclpubformat % this "if" is set by the choice of options

\else

\fi

\usepackage{xcolor} % Colored text
\usepackage{booktabs} % Table styles
\usepackage{pgfplots} % Make graphs from data
\usepackage{pgffor} % Foreach loops
\pgfkeys{/pgf/number format/1000 sep={\,}} % Change thousands separator in pgfplots
\usepackage{multicol} % Required for table (columns)
\usepackage{multirow} % Required for table (rows)
\usepackage{makecell} % Required for table (cells)
\usepackage{float} % Required for figure/table positioning
\usepackage{subfigure} 
\usepackage{arydshln}

\usepackage{graphicx}
\usepackage{tabularx}
\usepackage{lingmacros}
\newcommand{\suff}[1]{\textit{\textcolor{blue}{#1}}}

\newcommand{\textitbf}[1]{\textit{\textbf{#1}}}

\definecolor{baseline}{rgb}{0.95, 0.94, 0.94}
\definecolor{colfix1}{rgb}{0.82, 0.94, 0.75} %teagreen
\definecolor{colfix}{rgb}{0.67, 0.88, 0.69} % celadon
\definecolor{colrand}{rgb}{0.98, 0.81, 0.69} % apricot
\definecolor{colcase}{rgb}{0.8, 0.8, 1.0} %

\usepackage{color, colortbl}

\title{
On the Difficulty of Translating Free-Order Case-Marking Languages 
}

\author{Arianna Bisazza \qquad  Ahmet \"{U}st\"{u}n \qquad Stephan Sportel  \vspace{.2cm}
 \\ Center for Language and Cognition \\ 
 University of Groningen \vspace{.1cm}
 \\ {\sf \small\{a.bisazza, a.ustun\}@rug.nl, research@spor.tel}
}
\date{}

\begin{document}
\maketitle

\begin{abstract}
Identifying factors that make certain languages harder to model than others is essential to reach language equality in future Natural Language Processing technologies.
Free-order case-marking languages, such as Russian, Latin or Tamil, have proved more challenging than fixed-order languages for the tasks of syntactic parsing and subject-verb agreement prediction.
In this work, we investigate whether this class of languages is also more difficult to translate by state-of-the-art Neural Machine Translation models (NMT). 
Using a variety of synthetic languages and a newly introduced translation challenge set, we find that word order flexibility in the source language only leads to a very small loss of NMT quality, even though the core verb arguments become impossible to disambiguate in sentences without semantic cues.
The latter issue is indeed solved by the addition of case marking. However, in medium- and low-resource settings,  % case marking has a negative affect on MT quality in low-resource settings.
the overall NMT quality of \textit{fixed}-order languages remains unmatched. 
\end{abstract}

%\AB{Check/cite this for alternatives to MCC? A Comparison Between Morphological Complexity Measures: Typological Data vs. Language Corpora Christian Bentz} 

\section{Introduction}

%\AB{cite Dieuwke's JAIR about need for artificial data}
%\AB{check this: On the Computational Power of Transformers and Its Implications in Sequence Modeling Satwik Bhattamishra, Arkil Patel and Navin Goyal}

Despite the tremendous advances achieved in less than a decade, Natural Language Processing remains a field where language equality is far from being reached \cite{joshi-etal-2020-state}.
%Advances in neural network based models have led to astonishing improvements in machine translation \cite{sutskever2014sequence,bahdanau2015neural,vaswani2017attention} and represent the current state of the art in this task \cite{wu2016google}. 
In the field of Machine Translation, modern neural models %Neural Machine Translation (NMT) has become remarkable 
have attained remarkable quality
for high-resource language pairs like German-English, Chinese-English or English-Czech, with a number of studies claiming even human parity \cite{hassan2018achieving,bojar-etal-2018-findings,barrault-etal-2019-findings,Popel2020}.
These results may lead to the unfounded belief that NMT methods will perform equally well in any language pair, %are language-agnostic
%language independence has been reached 
provided similar amounts of training data. % in every language pair.
%The amount of available training data is of course a major explaining factor, 
In fact, several studies suggest the opposite
%even when data size is comparable, NMT accuracy can still vary dramatically among different language pairs 
\cite{platanios2018contextual,ataman2018evaluation,bugliarello-etal-2020-easier}.

Why, then, do some language pairs have lower translation accuracy? And, more specifically: Are certain typological profiles more challenging for current state-of-the-art NMT models? % and their intrinsic biases?
Every language has its own combination of typological properties, including word order, morphosyntactic features and more \cite{wals}. Identifying language properties (or combinations thereof) that pose major problems to the current modeling paradigms is essential to reach language equality in future MT (and other NLP) technologies \cite{joshi-etal-2020-state},
%design future MT (and more in general NLP) technologies that are truly language reach language independence in the design of future MT systems, 
in a way that is orthogonal to data collection efforts.
Among others, natural languages adopt different mechanisms to disambiguate the role of their constituents: Flexible order typically correlates with the presence of case marking and, vice versa, fixed order is observed in languages with little or no case marking \cite{comrie1981,Sinnemaki-tradeoff,futrell-15-quantifying}.
Morphologically rich languages \textit{in general} are known to be challenging for MT at least since the times of phrase-based statistical MT \cite{birch-etal-2008-predicting} due to their larger and sparser vocabularies, and remain challenging even for modern neural architectures \cite{ataman2018evaluation,belinkov-2017}. 
By contrast, the relation between word order flexibility and MT quality has not been directly studied to our knowledge.

In this paper, we study this relationship %by focusing on word order flexibilty in the source language, 
using strictly controlled experimental setups. Specifically, we ask:
\begin{itemize}
    \item Are current state-of-the-art NMT systems biased towards \textit{fixed-order} languages?
    \item To what extent does \textit{case marking} compensate for the lack of a fixed order in the source language?
\end{itemize}

%\AB{USE THIS HYPO:  Is it easier for an NMT model to learn the role of a word from its position within a sentence rather than from its case marking, even when other factors like vocabulary size and morphological ambiguity are controlled for? }

Unfortunately parallel data is scarce in most of the world languages \cite{guzman2019flores}, and corpora in different languages are drawn from different domains. 
Exceptions exist, like the widely used Europarl \cite{koehn2005europarl}, but represent a small fraction of the large variety of typological feature combinations attested in the world. % language spectrum. 
%
%Moreover, even if such a comparable corpus existed, isolating the impact of specific typological features would remain a challenge given that languages always differ from each other on many levels. 
This makes it very difficult to run a large-scale comparative study and isolate the factors of interest from, e.g., domain mismatch effects.
As a solution, we propose to evaluate NMT on synthetic languages \cite{gulordava-merlo-2016-multi,wang-eisner-2016-galactic,Ravfogel2019} that differ from each other only by specific properties, namely: the order of main constituents, or the presence and nature of case markers (see example in Table~\ref{tab:small-example}). 
%All variations are introduced in the source language, while the target language remains fixed.

We use this approach to isolate the impact of various source-language typological features on MT quality and to remove the typical confounders of corpus size and domain.
Using a variety of synthetic languages and a newly introduced challenge set, 
we find that state-of-the-art NMT has little to no bias towards fixed-order languages, but only when a sizeable training set is available.% in a high-resource setting, but 

\section{Free-order Case-marking Languages}

%Languages that primarily rely on the position of a word to encode its syntactic role have a typically rigid word order (e.g. English or Mandarin Chinese), while languages that rely on case-marking are often more flexible allowing for discourse-related factors like topicalization to be expressed through word order (e.g. Russian, Latin and Turkish).
%In this paper, we focus on the universal tendency of languages to trade-off between morphological complexity (particularly, case marking) and word order as a means to encode the role of their constituents \cite{comrie1981,Sinnemaki-tradeoff}.
%Flexible order typically correlates with the presence of case marking (e.g. Russian, Latin and Turkish) and, vice versa, fixed order is usually observed in languages with little or no case marking (e.g. English or Mandarin Chinese).  
The word order profile of a language is usually represented by the canonical order of its main constituents, (S)ubject, (O)bject, (V)erb.
For instance, English and French are SVO languages, while Turkish and Hindi are SOV. Other, less commonly attested, word orders are VSO and VOS, while OSV and OVS are extremely rare \cite{wals-81}. %The distribution of main word order types in a large sample of world languages is given in Table 3. 
While many other word order features exist (e.g., noun/adjective), they often correlate with the order of main constituents \cite{greenberg1963universals}.

A different, but likewise important dimension is that of word order \textit{freedom} (or \textit{flexibility}). Languages that primarily rely on the position of a word to encode grammatical roles typically display rigid orders (like English or Mandarin Chinese), while languages that rely on case marking can be more flexible allowing word order to express discourse-related factors like topicalization. 
Examples of highly flexible-order languages include languages as diverse as Russian, Hungarian, Latin, Tamil and Turkish.\footnote{See \newcite{futrell-15-quantifying} for detailed figures of word order freedom (measured by the entropy of subject and object dependency relation order) in a diverse sample of 34 languages.} 

In the field of psycholinguistics, due to the historical influence of English-centered studies, word order has long been considered the primary and most natural device through which children learn to infer syntactic relationships in their language \cite{slobin1966}.
However, cross-linguistic studies have later revealed that children are equally prepared to acquire both fixed-order and inflectional languages \cite{slobin1982}. % In fact, in the earliest developmental stages, learning to recognize case marking in a language like Turkish may even be \textit{easier} than paying attention to word order in a language like English \cite{slobin1982}.
%These interesting results were also replicated by simple recurrent networks (Elman) trained to detect the grammatical roles of words in small artificial languages (Lupyan).

Coming to computational linguistics, data-driven MT and other NLP approaches were also historically developed around languages with remarkably fixed order and very simple to moderately simple morphological systems, like English or French.
Luckily, our community has been giving increasing attention to more and more languages with diverse typologies, especially in the last decade.
So far, previous work has found that free-order languages 
%(i.e. with longer dependencies and non-projective dependecies) \cite{nivre-,mcdonals}
are more challenging for parsing \cite{gulordava-merlo-2015-diachronic,gulordava-merlo-2016-multi} and subject-verb agreement prediction \cite{Ravfogel2019} than their fixed-order counterparts.
This raises the question of whether word order flexibility also negatively affects MT quality.

\begin{table}[t]
\centering \small
\begin{tabular}{@{}l @{} c  @{\ }l@{}}
\toprule
\multirow{2}{*}{Fixed} & \sc vso & follows \underline{the little cat} the friendly dog \\
%\midrule
%\cline{2-3} 
 & \sc vos & follows  the friendly dog \underline{the little cat} \\
\midrule
 &  & follows \underline{the little cat\#S} the friendly dog\#O \\
\multicolumn{2}{@{}l}{Free+Case} & \multicolumn{1}{c}{OR} \\
& &  follows the friendly dog\#O \underline{the little cat\#S} \\
\midrule
\multicolumn{2}{@{}l}{Translation} & \underline{de kleine kat} volgt de vriendelijke hond \\
\bottomrule
\end{tabular}
\caption{Example sentence in different fixed/flexible-order English-based synthetic languages and their SVO Dutch translation. The subject in each sentence is \underline{underlined}. Artificial case markers start with \#.}
\label{tab:small-example}
\end{table}

Before the advent of modern NMT, \newcite{birch-etal-2008-predicting} used the Europarl corpus to study how various language properties affected the quality of phrase-based Statistical MT.
Amount of reordering, target morphological complexity, and historical relatedness of source and target languages were identified as strong predictors of MT quality.
% BIRCH: This paper investigates the effect of different ex- planatory variables on the performance of a phrase-based system for 110 European lan- guage pairs. We show that three factors are strong predictors of performance in isolation: the amount of reordering, the morphological complexity of the target language and the his- torical relatedness of the two languages. To- gether, these factors contribute 75% to the variability of the performance of the system
%
Recent work by \newcite{bugliarello-etal-2020-easier}, however,
has failed to show a correlation between NMT difficulty (measured by a novel information-theoretic metric) and several linguistic properties of source and target language, including Morphological Counting Complexity \cite{sagot-13} and Average Dependency Length \cite{Futrell10336}.
While that work specifically aimed at ensuring cross-linguistic comparability, the sample on which the linguistic properties could be computed (Europarl) was rather small and not very typologically diverse, leaving our research questions open to further investigation. 
In this paper, we therefore opt for a different methodology: namely, synthetic languages.

\section{Methodology}

\paragraph{Synthetic languages}
This paper presents two sets of experiments:
%To answer these questions we adopt two different methodologies. 
%
In the first (\S\ref{sect:expStephan}), we create parallel corpora using very simple and predictable artificial grammars and small vocabularies \cite{lupyan2002case}. See example in Table~\ref{tab:small-example}. By varying the position of subject/verb/object and introducing case markers to the source language, 
we study the biases of two NMT architectures in optimal training data conditions and a fully controlled setup, i.e. without any other linguistic cues that may disambiguate constituent roles.
In the second set of experiments (\S\ref{sect:expSynth}), we move to a more realistic setup using synthetic versions of the English language that differ from it in only one or few selected typological features \cite{Ravfogel2019}. For instance, the original sentence's order (SVO) is transformed to different orders, like SOV or VSO, based on its syntactic parse tree.

In both cases, typological variations are introduced in the source side of the parallel corpora, while the target language remains fixed. % (namely SVO without cases in the first experiment, French in the second experiment). 
In this way, we avoid the issue of non-comparable BLEU scores across different target languages. 
% NOW IN CONCLUSION: \footnote{Extending our study to different target languages using cross-mutual information \cite{bugliarello-etal-2020-easier} instead of BLEU is an interesting avenue for future work.}
%
Lastly, we make the simplifying assumption that, when verb-argument order varies from the canonical order in a flexible-order language, it does so in a totally arbitrary way.
While this is rarely true in practice, %we believe this assumption should not affect the generality 
as word order may be predictable given pragmatics or other factors, we focus here on \textit{``the extent to which word order is conditioned on the syntactic and compositional semantic properties of an utterance''} \cite{futrell-15-quantifying}.

\paragraph{Translation models}
We consider two widely used NMT architectures that crucially differ in their encoding of positional information: (i) Recurrent sequence-to-sequence BiLSTM with attention \cite{bahdanau2015neural,Luong2015} processes the input symbols sequentially and has each hidden state directly conditioned on that of the previous (or following, for the backward LSTM) timestep \cite{elman1990,hochreiter1997long}. 
(ii) The non-recurrent, fully attention-based Transformer  \cite{Vaswani2017} processes all input symbols in parallel relying on dedicated embeddings to encode each input's position.\footnote{We use sinusoidal embeddings \cite{Vaswani2017}. All our models are built using OpenNMT: \url{https://github.com/OpenNMT/OpenNMT-py}}
Transformer has nowadays surpassed recurrent encoder-decoder models in terms of generic MT quality. 
Moreover, \newcite{choshen-abend-2019-automatically} have recently shown that Transformer-based NMT models are indifferent to the absolute order of source words, at least when equipped with learned positional embeddings. On the other hand, the lack of recurrence in  Transformers has been linked to a limited ability to capture hierarchical structure \cite{tran-etal-2018-importance,hahn19-transf-limit}. 
To our knowledge, no previous work has studied the biases of either architectures towards fixed-order languages in a systematic manner.

\section{Toy Parallel Grammar}
\label{sect:expStephan}

%\AB{To DECIDE: Leave this here or move down as an extra set of exps?}

%In this first experiment we make use of three self-made parallel corpora.
%We do this to ensure a controlled environment for our research.
%Because languages vary on many levels, such as word order and morphological system, using an existing corpus would make it difficult to isolate the language properties we are researching.

We start by evaluating our models on a pair of toy languages inspired by the English-Dutch pair and created using a Synchronous Context-Free Grammar \citep{Chiang2006}.
Each sentence consists of a simple clause with a transitive verb, subject and object. Both arguments are singular and optionally modified by an adjective.
The source vocabulary contains 6 nouns, 6 verbs, 6 adjectives, % and a determiner (19 types), 
and the complete corpus contains 10k generated sentence pairs. %\footnote{Sentences containing the same noun twice are excluded.}
Working with such a small, finite grammar allows us to simulate an otherwise impossible situation where the NMT model can be trained on (almost) the totality of a language's utterances, canceling out data sparsity effects.\footnote{Data and code to replicate the toy grammar experiments in this section are available at \url{https://github.com/573phn/cm-vs-wo}}

%Our experiment is reminiscent of Lupyan, with some notable differences:  case marking representations were hard-coded and not learned from data, their language design was more ... the task was grammatical role classification and not translation to another language.

\begin{figure*}[ht]
\centering
\subfigure[BiLSTM with attention]{
    \includegraphics[width=.69\columnwidth, keepaspectratio]{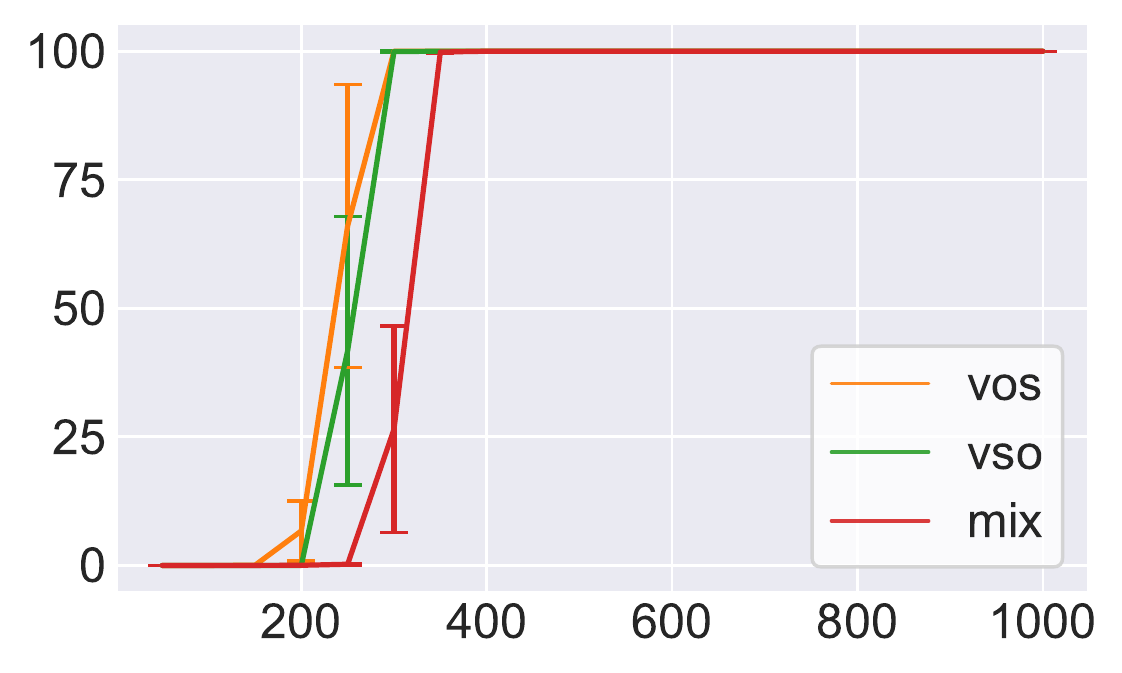}    
    \label{fig:with}
} \hspace{-0.5cm}
\subfigure[Large Transformer]{
    \includegraphics[width=.69\columnwidth, keepaspectratio]{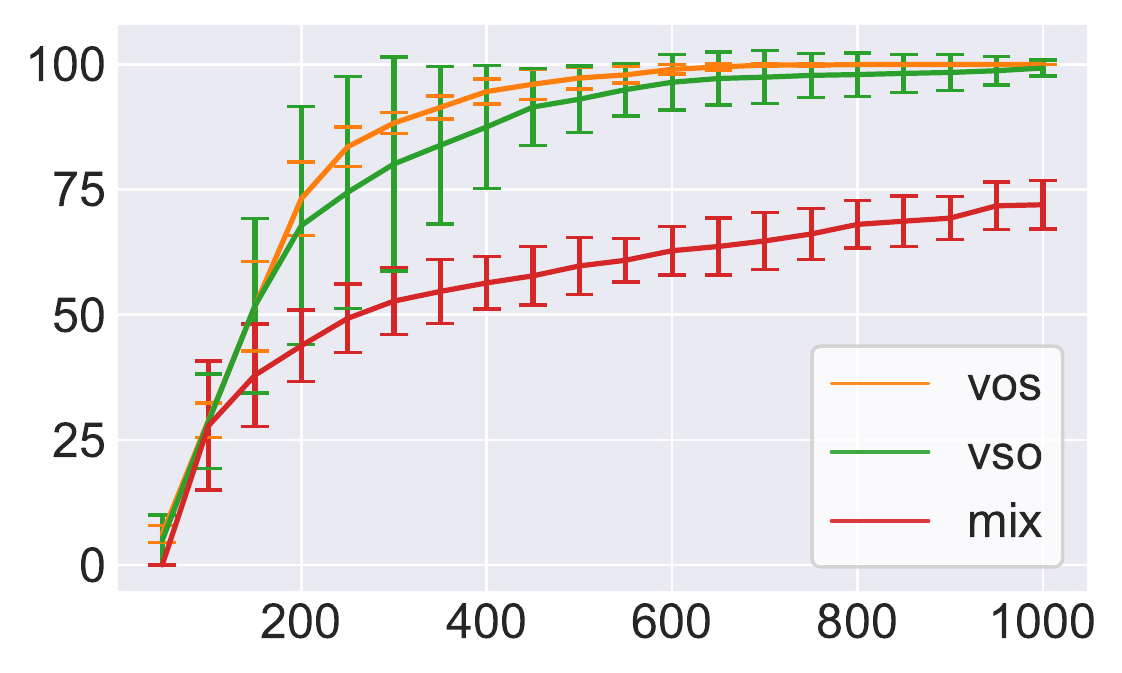}
    \label{fig:largesgd}
} \hspace{-0.5cm}
\subfigure[Small Transformer]{
   \includegraphics[width=.69\columnwidth, keepaspectratio]{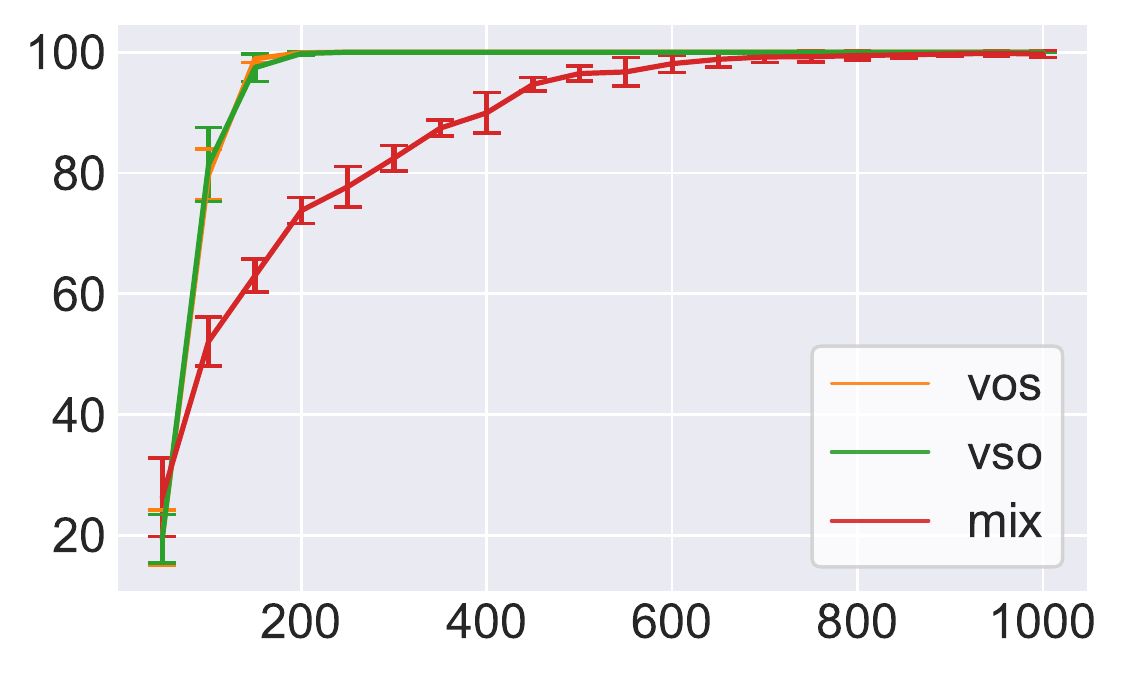}
    \label{fig:smallsgd}
}
\caption{Toy language NMT sentence-level accuracy on validation set by number of training epochs. Source languages: fixed-order VSO, fixed-order VOS, and mixed-order (VSO/VOS) with case marking.
%NMT learning curves (sentence-level accuracy) of NMT trained on three different source toy languages: fixed-order VSO, fixed-order VOS, and mixed-order (VSO/VOS) with case marking.
Target language: always fixed SVO. Each experiment is repeated five times, and averaged results are shown.} %with error bars for standard deviation.}
\label{fig:toyResults}
\end{figure*}

\paragraph{Source Language Variants}
We consider three source language variants, illustrated in Table~\ref{tab:small-example}:\begin{itemize}
\item fixed-order VSO;
\item fixed-order VOS; 
\item mixed-order (randomly chosen between VSO or VOS) with nominal case marking. 
\end{itemize}
We choose these word orders so that, in the flexible-order corpus, % the location respective to the verb does not reveal which noun is the subject and which is the object. As a result,
the only way to disambiguate argument roles is case marking, realized by simple unambiguous suffixes (\textit{\#S} and \textit{\#O}). 
The target language is always fixed SVO.
% VOCAB SIZES:
%The vocabularies of the VOS and VSO corpora each contain 19 unique words in the source language, while this number is 25 for the vocabulary of the mixed corpus.The mixed corpus contains six more unique words than the VOS and VSO corpora because each of the six nouns is present with both an object and subject suffix.
%
The same random split (80/10/10\% training/validation/test) is applied to the three corpora.

\paragraph{NMT Setup}

As recurrent model, we trained a 2-layer BiLSTM with attention \citep{Luong2015} with 500 hidden layer size.
As Transformer models, we trained one using the standard 6-layer configuration \cite{Vaswani2017} and a smaller one with only 2 layers given the simplicity of the languages. 
%Both Transformer architectures use sinusoidal position encoding \cite{TODO}.
All models are trained at the word level using the complete vocabulary.
More hyper-parameters are provided in Appendix~\ref{app:hparams}.
Note that our goal is not to compare LSTM and Transformer accuracy to each other, but rather to observe the different trends across fixed- and flexible-order language variants.
Given the small vocabulary, we use sentence-level accuracy instead of BLEU for evaluation.

\paragraph{Results} % +discussion
\label{sect:resultsToy}

As shown in Figure~\ref{fig:toyResults}, all models achieve perfect accuracy on all language pairs after 1000 training steps, except for the Large Transformer on the free-order language, likely due to overparametrization \cite{Sankararaman20-overparam}.
These results demonstrate that our NMT architectures are equally capable of modeling translation of both types of language, when all other factors of variation are controlled for.
Nonetheless, a pattern emerges when looking at the learning curves within each plot: While the two fixed-order languages have very similar learning curves,
the free-order language with case markers always requires slightly more training steps to converge. 
This is also the case, albeit to a lesser extent, when the mixed-order corpus is pre-processed by splitting all case suffixes from the nouns (extra experiment not shown in the plot).
This trend is noteworthy, given the simplicity of our grammars and the transparency of the case system.
As our training sets cover a large majority of the languages, % (except for the 20\% sentences reserved for validation and test), 
this result might suggest that free-order \textit{natural} languages need larger training datasets to reach a similar translation quality than their fixed-order counterparts. In \S{\ref{sect:expSynth}} we validate this hypothesis on more naturalistic language data.

%\AB{VERIFY THE ABOVE how many steps to process the whole training set?}

\section{Synthetic English Variants}
\label{sect:expSynth}

Experimenting with toy languages has its shortcomings, like the small vocabulary size and non-realistic distribution of words and structures.
In this section, we follow the approach of \newcite{Ravfogel2019} to validate our findings in a less controlled but more realistic setup.
Specifically, we create several variants of the Europarl English-French parallel corpus where the source sentences are modified by changing word order and adding artificial case markers.
We choose French as target language because of its fixed order, SVO, and its relatively simple morphology.\footnote{According to the Morphological Counting Complexity \cite{sagot-13} values reported by \newcite{cotterell-etal-2018}, English scores 6 (least complex), Dutch 26, French 30, Spanish 71, Czech 195, and Finnish 198 (most complex).}
As Indo-European languages, English and French are moderately related in terms of syntax and vocabulary while being sufficiently distant to avoid a word-by-word translation strategy in many cases.

Source language variants are obtained by transforming the syntactic tree of the original sentences.
While \newcite{Ravfogel2019} could rely on the Penn Treebank \cite{marcus1993building} for their monolingual task of agreement prediction, we instead need parallel data. For this reason, we parse the English side of the Europarl~v.7 corpus \cite{koehn2005europarl} %\footnote{\url{https://www.statmt.org/europarl/v7}} 
using the Stanza dependency parser \cite{qi2020stanza, manning2014}.
After parsing, we adopt a modified version of the synthetic language generator by \newcite{Ravfogel2019} to create the following English variants:\footnote{%\url{https://github.com/Shaul1321/rnn typology},
Our revised language generator is available at \url{https://github.com/573phn/rnn_typology}}
\begin{itemize}
    \item \textbf{fixed-order}: either SVO, SOV, VSO or VOS;\footnote{To keep the number of experiments manageable, we omit object-initial languages which are significantly less attested among world languages \cite{wals-81}.}
    \item \textbf{free-order}: for each sentence in the corpus, one of the six possible orders of (Subject, Object, Verb) is chosen randomly;
    %\item \textbf{partly flexible order}: as most natural languages display at least \textit{some} degree of flexibility \cite{futrell-15-quantifying}, we also experiment with language variants where one of the six orders is dominant, i.e. selected in 30\% or 60\% of the sentences, and a different order is randomly selected in the remaining sentences;
    \item \textbf{shuffled words}: all source words are shuffled regardless of their syntactic role. This is our lower bound, measuring the reordering ability of a model in the total absence of source-side order cues (akin to bag-of-words input).
\end{itemize}
To allow for a fair comparison with the artificial case-marking languages, 
we remove number agreement features from verbs in all the above variants
% subject-verb agreement is removed from all the above variants,
%by removing number marking from verbs 
(cf. \textit{says} $\rightarrow$ \textit{say} in Table~\ref{tab:cases}).

% From Ravfogel:
% In the typological survey reported in Baerman and Brown (2013), 62% of the languages had no or minimal case marking, 20% had syncretic case systems, and 18% had case systems with no syn- cretism.

To answer our second research question, we experiment with two artificial case systems proposed by \newcite{Ravfogel2019} and illustrated in Table~\ref{tab:cases} (overt suffixes):
\begin{itemize}
    \item \textbf{unambiguous case system}: %artificial 
    suffixes indicating argument role (subject/object/indirect object) and number (singular/plural) are added to the heads of noun and verb phrases; %after removing possibly existing English suffixes;
    \item \textbf{syncretic case system}: suffixes indicating number but not grammatical function are added to the heads of main arguments, providing only partial disambiguation of argument roles. This system is inspired from subject/object syncretism in Russian.
    %A fully syncretic case system (argument marking only): the suffix indicated only the plurality of the argument, regardless of its grammatical function (cf. subject/object syn- cretism in Russian neuter nouns
\end{itemize}
Syncretic case systems were found to be roughly as common as non-syncretic ones in a large sample of almost 200 world languages \cite{wals-28-caseSyncretism}.
Case marking is always combined with the fully flexible order of main constituents.
As in \cite{Ravfogel2019}, English number marking is removed from verbs and their arguments before adding the artificial suffixes. 

\begin{table}[t]
\begin{small}
{\renewcommand{\arraystretch}{1.1}
\begin{tabular}{@{\ }p{7.75cm}@{}}
\hline
\multicolumn{1}{@{\ }l}{\textitbf{Original (no case):}}\\
%The woman says her sisters took her out for dinner frequently.\\
The woman says her sisters often invited her for dinner.\\
\hline
\multicolumn{1}{@{\ }l}{\textitbf{SOV (no case):}} \\
The woman her sisters her often invited for dinner say. \\
\hline
\multicolumn{1}{@{\ }l}{\textitbf{SOV, syncretic case marking (overt):}} \\
The woman\suff{.arg.sg} her sisters\suff{.arg.pl} she\suff{.arg.sg} often invited\suff{.arg.pl} for dinner say\suff{.arg.sg}. \\
\hline
\multicolumn{1}{@{\ }l}{\textitbf{SOV, unambiguous case marking (overt):}} \\
The woman\suff{.nsubj.sg} her sisters\suff{.nsubj.pl} she\suff{.dobj.sg} often invited\suff{.dobj.sg.nsubj.pl} for dinner say\suff{.nsubj.sg}. \\
\hline
\multicolumn{1}{@{\ }l}{\textitbf{SOV, unambiguous case (implicit):}} \\
The woman\suff{kar} her sisters\suff{kon} she\suff{kin} often invited\suff{kinkon} for dinner say\suff{kar}. \\
\hline
\multicolumn{1}{@{\ }l}{\textitbf{SOV, unambiguous case (implicit with declensions):}} \\
The woman\suff{kar} her sisters\suff{pon} she\suff{kit} often invited\suff{kitpon} for dinner say\suff{kar}. \\
\hline
\multicolumn{1}{@{\ }l}{\textitbf{French translation:}} \\
La femme dit que ses soeurs l'invitaient souvent à dîner.\\
\hline
\end{tabular}
}
\end{small}
\caption{Examples of synthetic English variants and their (common) French translation. The full list of suffixes is provided in Appendix~\ref{app:suffixes}.}
\label{tab:cases}
\end{table}

% DECLENSIONS
% woman 1
% sister 2
% she 3

% TODO ITALIAN translation
% l'homme a dit que sa femme l'emmenait souvent dîner.
% l'uomo disse che sua moglie lo portava spesso fuori a cena

\subsection{NMT Setup}

\paragraph{Models}
As recurrent model, we used a 3-layer BiLSTM with hidden size of 512 and MLP attention \cite{bahdanau2015neural}. The Transformer model has the standard 6-layer configuration with hidden size of 512, 8 attention heads, and sinusoidal positional encoding \cite{Vaswani2017}.
All models use subword representation based on 32k BPE merge operations \cite{sennrich-etal-2016-neural}, except in the low-resource setup where this is reduced to 10k operations. 
More hyper-parameters are provided in Appendix~\ref{app:hparams}.

\paragraph{Data and Evaluation}
We train our models on various subsets of the English-French Europarl corpus:
1,9M sentence pairs (high-resource), 100K (medium-resource), 10K (low-resource).
For evaluation, we use 5K sentences randomly held-out from the same corpus. Given the importance of word order to assess the correct translation of verb arguments into French, we compute the reordering-focused RIBES\footnote{BLEU captures local word-order errors only indirectly (lower precision of higher-order n-grams) and does not capture long-range word-order errors at all. By contrast, RIBES directly measures correlation between the word ranks in the reference and those in the MT output.} 
metric \cite{ribes-isozaki} in addition to the more commonly used BLEU \cite{Papineni2002}.
% RIBES measures the correlation of n-gram ranks between the output and the reference, where n-gram appears uniquely and in both.
% Our method is based on rank correlation co- efficients. We use them to compare the word ranks in the reference with those in the hypothesis
In each experiment, the source side of training and test data is transformed using the same procedure whereas the target side remains unchanged.
We repeat each experiment 3 times (or 4 for languages with random order choice) and report the averaged results.

\subsection{Challenge Set}
\label{sect:challenge}

Besides syntactic structure, natural language often contains semantic and collocational cues that help disambiguate the role of an argument. Small BLEU/RIBES differences between our language variants may indicate actual robustness of a model to word order flexibility, 
but may also indicate that a model relies on those cues rather than on syntactic structure \cite{gulordava-etal-2018-colorless}. To discern these two hypotheses, we create a challenge set of 7,200 simple affirmative and negative sentences where swapping subject and object leads to another plausible sentence.\footnote{More details can be found in Appendix~\ref{app:challenge}. We release the challenge set at \url{https://github.com/arianna-bis/freeorder-mt}} 
Each English sentence and its reverse are included in the test set together with the respective translations, as for example:
%For example, the original (SVO) version of the set includes:
\eenumsentence{
\item The president thanks the minister. /\\ Le président remercie le ministre.
\item The minister thanks the president. /\\ Le ministre remercie le président.
}
The source side is then processed as explained in \S\ref{sect:expSynth} and translated by the NMT model trained on the corresponding language variant.
%This way, perfect accuracy can only be obtained by a model properly taking into account syntactic structure.
Thus, translation quality on this set reflects the extent to which NMT models have robustly learnt to detect verb arguments and their roles independently from other cues, which we consider an important sign of linguistic generalization ability. 
%This set also allows us to draw a comparison to the results reported by \newcite{Ravfogel2019} on subject-verb agreement.
%
For space constraints we only present RIBES scores on the challenge set.\footnote{% 
%we are primarily interested in the model's ability to correctly translate the sentence structure.
We also computed BLEU scores: they strongly correlate with RIBES but fluctuate more due to the larger effect of lexical choice.}
%\AB{TODO appendix: BLEU on challenge set. NOT NEEDED?}

%In preliminary experiments we found small BLEU and RIBES differences among language variants. This may be due to an actual robustness of NMT models to word order flexibility, but also to the relatively small impact of verb argument position on the overall sentence

%%%%%
%%%%%
%\input{main-results-OLDmonoLSTM.tex}
\begin{table*}[ht]
%\small
\centering
\begin{tabular}{@{\ }l @{\hskip 0.26in} l l l @{\hskip 0.2in}lll@{\ }}
\toprule
English*$\rightarrow$French &  \multicolumn{3}{c}{\textsc{bi-lstm}\ \ \ }      & \multicolumn{3}{c}{\textsc{transformer}}        \\
Large Training (1.9M)           & \multicolumn{2}{l}{Europarl-Test} & Challenge & \multicolumn{2}{l}{Europarl-Test} & Challenge \\ 
\midrule
             & \textsc{bleu}     & \textsc{ribes}      & \textsc{ribes}             & \textsc{bleu}     & \textsc{ribes}        & \textsc{ribes}       \\ \midrule
\rowcolor{baseline}
Original English            & 39.4 & 85.0  & 98.0  & 38.3        & 84.9        & 97.7        \\
\midrule 
\multicolumn{7}{@{\ }l}{\textit{Fixed Order:}} \\ \midrule
\rowcolor{colfix1} 
S-V-O                        & 38.3 & 84.5  & 98.1  & 37.7        & 84.6        & 98.0        \\
\rowcolor{colfix1}
S-O-V                        & 37.6 & 84.2  & 97.7  & 37.9        & 84.5        & 97.2        \\
\rowcolor{colfix1}
V-S-O                        & 38.0 & 84.2  & 97.8  & 37.8        & 84.6        & 98.0        \\
\rowcolor{colfix1}
V-O-S                          & 37.8 & 84.0  & 98.0  & 37.6        & 84.3       & 97.2        \\
\rowcolor{colfix} 
Average (fixed orders) & 37.9$\pm$0.4 & 84.2$\pm$0.3  & 97.9$\pm$0.2  & 37.8$\pm$0.1        & 84.5$\pm$0.1        & 97.6$\pm$0.4        \\

\midrule
\multicolumn{7}{@{\ }l}{\textit{Flexible Order:}} \\ 
\midrule
% RESULTS OF DOMINANT ORDER LANGUAGES:
%V-S-O dominant (60\%)        & 35.9 & 83.4  & 62.6 & 83.7  & 37.7        & 84.4        & 65.4  & 82.4        \\
%V-O-S dominant (60\%)        & 35.9 & 83.2  & 63.1 & 83.1  & 37.4        & 84.2        & 68.1  & 83.8        \\
%V-S-O dominant (30\%)        & 35.6 & 83.0  & 51.6 & 76.4  & 37.6        & 84.2        & 55.6  & 74.8        \\
%V-O-S dominant (30\%)        & 35.5 & 83.0  & 50.6 & 75.6  & 37.6        & 84.1        & 57.4  & 75.2        \\
\rowcolor{colrand}
Random, no case                      & 37.1 & 83.7 & 75.1 & 37.5        & 84.2 & 74.1        \\
%\midrule
%\multicolumn{9}{l}{\textit{Flexible Order \& Case: }} \\ \midrule
\rowcolor{colcase} 
Random + syncretic case     & 36.9 & 83.6 & 75.4  & 37.3        & 84.2        & 84.4        \\
\rowcolor{colcase} 
Random + unambig. case   & 37.3 & 83.9  & 97.7  & 37.3        & 84.4       & 98.1        \\
%Random + unamb. fusional    & 35.8 & 83.3  & 79.8 & 97.3  & 37.3        & 84.3        & 82.8  & 97.5        \\
%Random + unamb. fus. decl.  & 35.8 & 83.4  & 83.9 & 98.0  & 37.4        & 84.3        & 83.1  & 97.6        \\
 \midrule
%\multicolumn{7}{@{\ }l}{\textit{Shuffle All:}} \\ 
%\midrule
 \rowcolor{baseline}
Shuffle all words           & 18.5 & 65.2  & 79.4  & 25.8        & 71.2        & 83.2        \\ 
\bottomrule
\end{tabular}
\caption{Translation quality from various English-based synthetic languages into standard French, using the largest training data (1.9M sentences).
NMT architectures: 3-layer BiLSTM seq-to-seq with attention; 6-layer Transformer.
Europarl-Test: 5K held-out Europarl sentences; Challenge set: see \S\ref{sect:challenge}. All scores are averaged over three training runs.
%\AB{TODO! MAKE TABLE FIT HORIZONTALLY!! }
}
\label{tab:main-results}
\end{table*}

%%%%%
%%%%%

\subsection{High-Resource Results}
\label{sect:highResResults}

% ABSTRACT:
% we find that word order flexibility by itself leads to only a very small drop of overall NMT quality, even though the core verb arguments become impossible to disambiguate in short sentences with no semantic cues. 
% The latter issue is indeed solved by the addition of case marking. However, the overall NMT quality of \textit{fixed}-order languages remains unmatched, possibly due to the increased data sparsity caused by case marking.

Table~\ref{tab:main-results} reports the high-resource setting results.
The first row (original English to French) is given only for reference and shows the overall highest results. The BLEU drop observed when moving to any of the fixed-order variants (including SVO) is likely due to parsing flaws resulting in awkward reorderings. As this issue affects all our synthetic variants, it does not undermine the validity of our findings. 
For clarity, we center our main discussion on the Transformer results and comment on the BiLSTM results at the end of this section.

\paragraph{Fixed-Order Variants}
All four tested fixed-order variants obtain very similar BLEU/RIBES scores on the Europarl-test. This is in line with previous work in NMT showing that linguistically motivated pre-ordering leads to small gains \cite{Zhao2018} or none at all \cite{Du2017}, and that Transformer-based models are \textit{not} biased towards monotonic translation \cite{choshen-abend-2019-automatically}.
On the challenge set, scores are slightly more variable but a manual inspection reveals that this is due to different lexical choices, while word order is always correct for this group of languages.
To sum up, in the high-resource setup, our Transformer models are perfectly able to disambiguate the core argument roles when these are consistently encoded by word order.

\paragraph{Fixed-Order vs Random-Order}
Somewhat surprisingly, the Transformer results are only marginally affected by the random ordering of verb and core arguments. Recall that in the `Random' language all six possible permutations of (S,V,O) are equally likely. 
Thus, Transformer shows an excellent ability to reconstruct the correct constituent order \textit{in the general-purpose test set}. 
The picture is very different on the challenge set, where RIBES drops severely from 97.6 to 74.1.
These low results were to be expected given the challenge set design (it is impossible even for a human to recognize subject from object in the `Random, no case' challenge set). Nonetheless, they demonstrate that the general-purpose set cannot tell us whether an NMT model has learnt to reliably exploit syntactic structure of the source language, because of the abundant non-syntactic cues.
In fact, even when \textit{all} source words are shuffled, Transformer still achieves a respectable  25.8/71.2 BLEU/RIBES on the Europarl-test.

%\paragraph{Random-Order vs Dominant-Order}
%\AB{MAYBE REMOVE THESE RESULTS AFTER ALL???? There's too little gap between Fixed- and Random-order for anything interesting to emerge here...}
%In natural languages, the distribution of constituent orders is rarely \textit{i.i.d.} 
%Would our results differ if one of the six possible orders were dominant, as is typically the case in the real world? 
%Our challenge set results show that the NMT models do not learn to  ...

% The latter issue is indeed solved by the addition of case marking. However, the overall NMT quality of \textit{fixed}-order languages remains unmatched, possibly due to the increased data sparsity caused by case marking.

\paragraph{Case Marking}
The key comparison in our study lies between fixed-order and free-order case-marking languages.
%The challenge set results show that case marking 
Here, we find that case marking can indeed restore near-perfect accuracy on the challenge set (98.1 RIBES). %thereby mirroring the results of our toy grammar experiments (\S\ref{sect:resultsToy}).
However, this only happens when the marking system is completely unambiguous, which, as already mentioned, is true for only about a half of the real case-marking languages \cite{wals-28-caseSyncretism}.
Indeed the syncretic system visibly improves quality on the challenge set (74.1 to 84.4 RIBES) but remains far behind the fixed-order score (97.6).
%However, the overall NMT quality of \textit{fixed}-order languages remains unmatched, possibly due to the increased data sparsity caused by case marking.
%
In terms of overall NMT quality (Europarl-test), fixed-order languages score only marginally higher than the free-order case-marking ones, regardless of the unambiguous/syncretic distinction.
Thus our finding that Transformer NMT systems are equally capable of modeling the two types of languages (\S\ref{sect:resultsToy}) 
is also confirmed with more naturalistic language data.
%is confirmed also in the more realistic setup of English-based synthetic languages.
%Thus, even in the more realistic setup of English-based synthetic languages, we confirm that Transformer-based NMT systems are equally capable of modeling the two types of languages.
%
That said, we will show in Sect.~\ref{sect:dataSize} that this positive finding is conditional on the availability of large amounts of training samples.

%we find that both case marking systems yield only a tiny BLEU/RIBES loss (from 37.5/84.2 to 37.3/84.4

\paragraph{BiLSTM vs Transformer} 
The LSTM-based results generally correlate with the Transformer results discussed above, however our recurrent models appear to be slightly more sensitive to changes in the source-side order, in line with previous findings \cite{choshen-abend-2019-automatically}. 
Specifically, translation quality on Europarl-test fluctuates slightly more than Transformer among different fixed orders, with the most monotonic order (SVO) leading to the best results. 
%, and (ii) drops slightly more when core arguments are randomly ordered.
% CAMERA-READY
%($-1.0$/$-0.8$ BLEU/RIBES \textit{vs} $-0.3$/$-0.3$ for Transformer);
When \textit{all} words are randomly shuffled,
BiLSTM scores drop much more than Transformer. % when all words are shuffled.
However, when comparing the fixed-order variants to the ones with free order of main constituents, % case-marking ones, 
BiLSTM shows only a slightly stronger preference for fixed-order, compared to Transformer. %, however the difference is smaller than one could have expected. 
%- larger variation among fixed-orders (on eparl, same on challenge)
%- larger BLEU/RIBES drop from avg fixed to random,no case (on eparl, same on challenge)
%- also shuffle much worse
%- key comparison: gap between fixed-order and rand+case a bit larger than in Transformer, may not be significant
This suggests that, by experimenting with arbitrary permutations, \newcite{choshen-abend-2019-automatically} might have overestimated the bias of recurrent NMT towards more monotonic translation,
whereas the more realistic combination of constituent-level reordering with case marking used in our study is not so problematic for this type of model.

Interestingly, on the challenge set, BiLSTM and Transformer perform on par, with the notable exception that syncretic case is much more difficult for the BiLSTM model.
%
%Our BiLSTM results on the challenge set
Our results agree with the large drop of subject-verb agreement prediction accuracy  observed by \newcite{Ravfogel2019} when experimenting with the random order of main constituents.
However, their scores were also low for SOV and VOS, which is not the case in our NMT experiments.
Besides the fact that our challenge set only contains short sentences (hence no long dependencies and few agreement attractors), our task is considerably different in that agreement only needs to be predicted in the target language, which is fixed-order SVO. 

%Beyond translation, our LSTM results contrast with the large variations in subject-verb agreement accuracy observed by \newcite{Ravfogel2019} when varying the order of main constituents. Indeed, their scores were particularly low for SOV, VOS and random, which is not the case in our experiments, not even on our targeted challenge set.

%\section{Effect of Data Size, Morphological Features, and Target Language}
%\label{sect:extraEffects}

%\vspace{1mm}
\paragraph{Summary}
Our results so far suggest that state-of-the-art NMT models, especially if Transformer-based, have little or no bias towards fixed-order languages. In what follows, we study whether this finding is robust to differences in data size, type of morphology, and target language.

\begin{figure*}[ht]
\centering
\subfigure[Europarl-Test BLEU]{
    \includegraphics[width=.655\columnwidth, keepaspectratio]{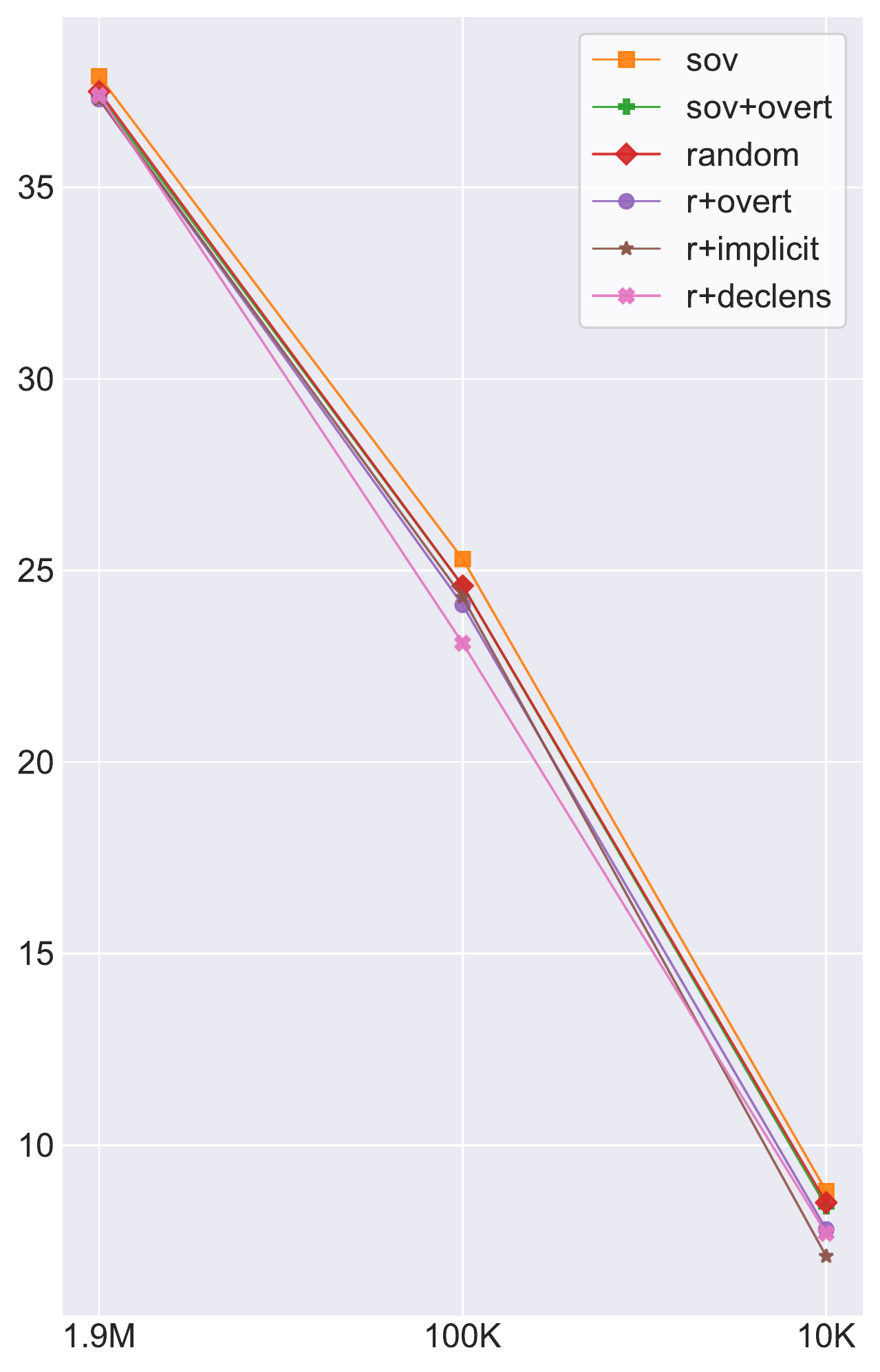}    
    \label{fig:eparl-bleu}
}
\subfigure[Europarl-Test RIBES]{
    \includegraphics[width=.655\columnwidth, keepaspectratio]{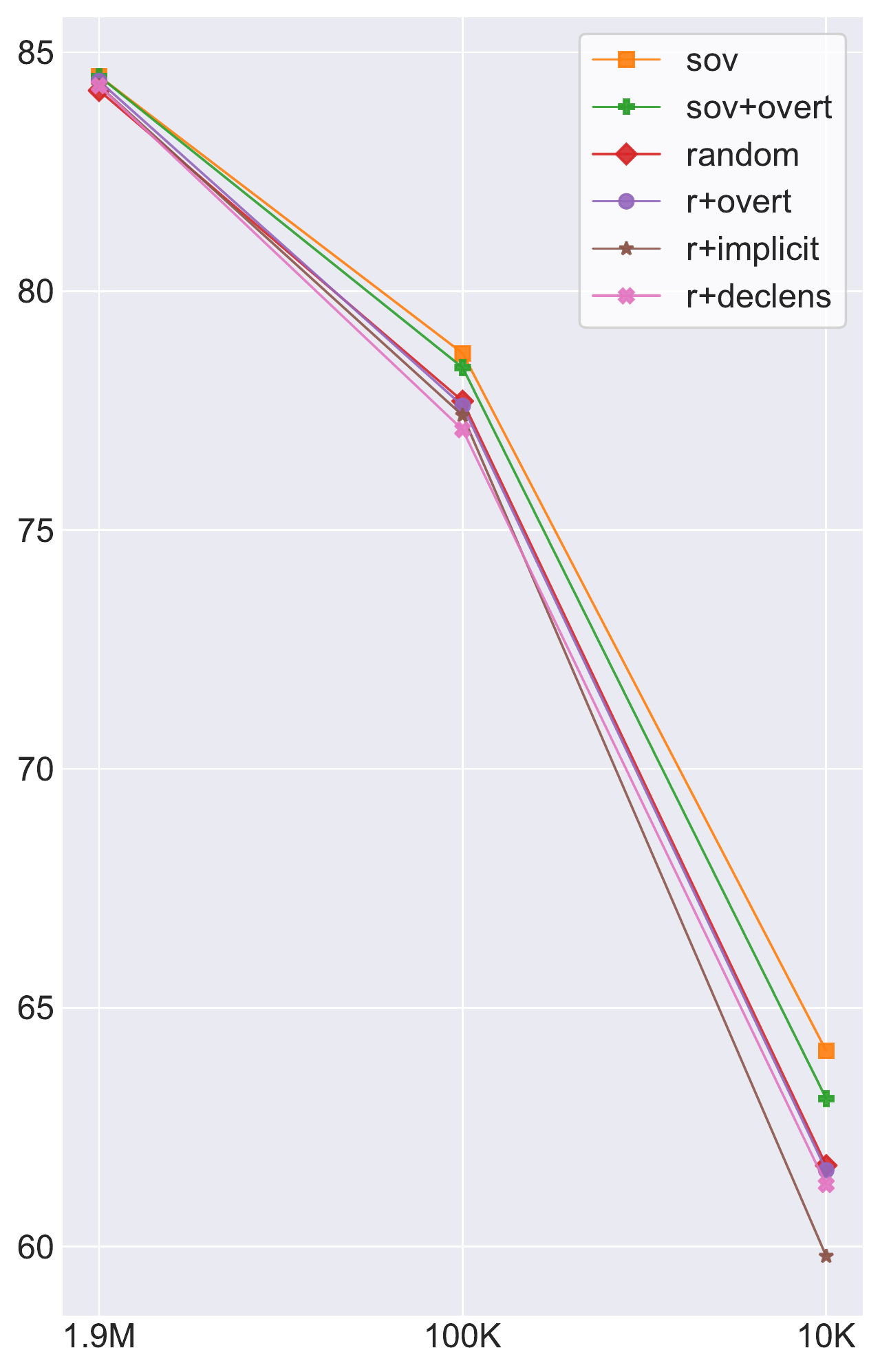} 
    \label{fig:eparl-ribes}
}
\subfigure[Challenge RIBES]{
    \includegraphics[width=.655\columnwidth, keepaspectratio]{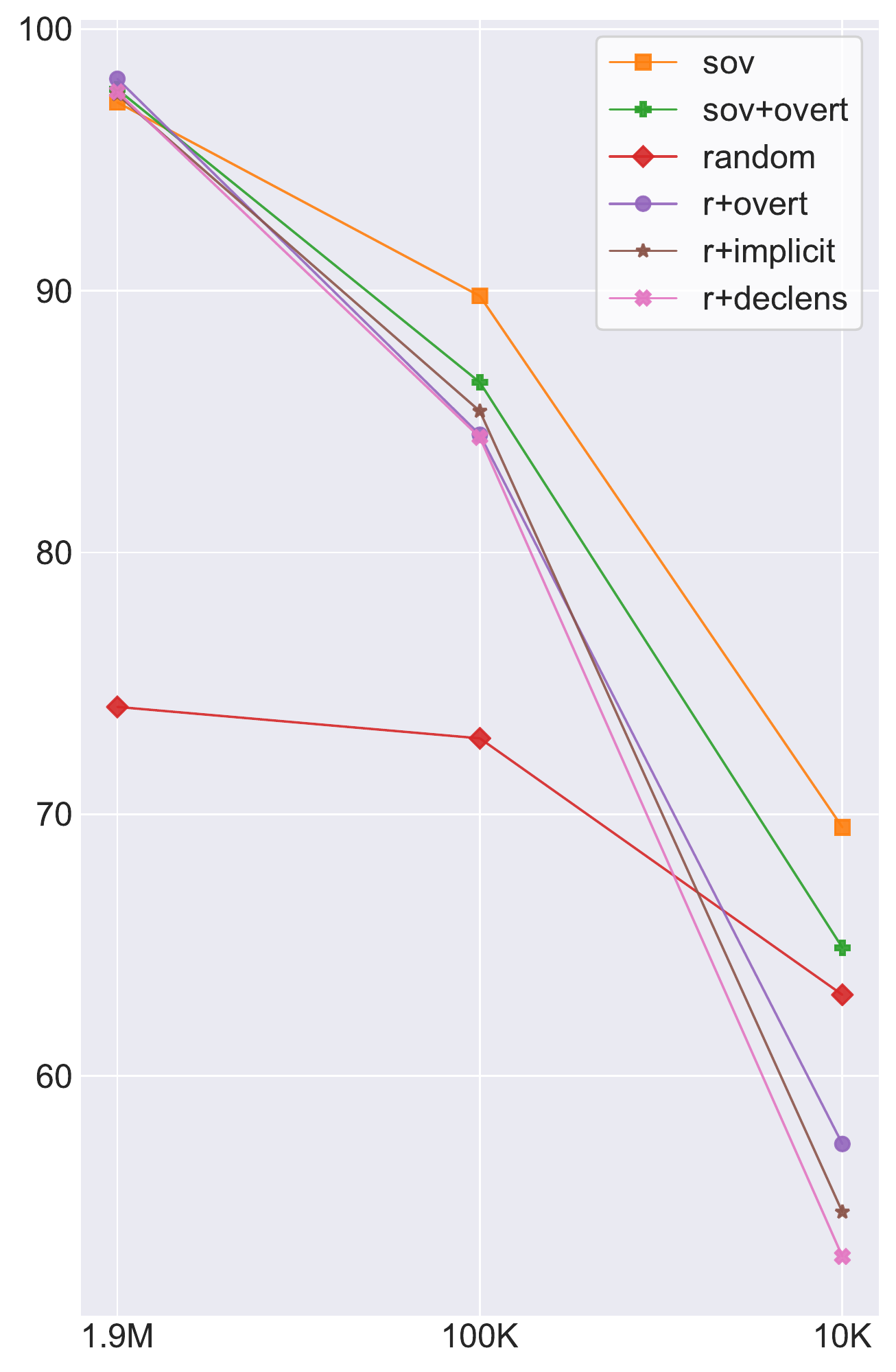}
    \label{fig:control-ribes}
}
\caption{EN*-FR Transformer NMT quality versus training data size (x-axis). Source language variants: Fixed-order (SOV) and free-order (random) with different case systems (r+overt/implicit/declens). Scores averaged over three training runs. Detailed numerical results are provided in Appendix~\ref{app:figure2}.}
\label{fig:datasize}
\end{figure*}

\subsection{Effect of Data Size and Morphological Features}
\label{sect:dataSize}

\paragraph{Data Size}
The results shown in Table~\ref{tab:main-results} represent a high-resource setting (almost 2M training sentences). %, however parallel data is scarce in most of the world languages \cite{guzman2019flores}.
While recent successes in cross-lingual transfer learning alleviate the need for labeled data \cite{liu-etal-2020-mbart}, % in the target language, but 
their success still depends on the availability of large unlabeled data as well as other, yet to be explained, language properties \cite{joshi-etal-2020-state}.
%Even though recent successes in cross-lingual transfer learning \AB{CITE} promise to alleviate the need for labeled data in the target language, many languages also suffer from the lack of unlabeled data and typology
We then ask: Do free-order case-marking languages need more data than fixed-order non-case-marking ones to reach similar NMT quality?
We simulate a medium- and low-resource scenario by sampling 100K and 10K training sentences, respectively, from the full Europarl data.
To reduce the number of experiments, we only consider Transformer with one fixed-order language variant (SOV)\footnote{%Given previous results \cite{choshen-abend-2019-automatically} and our own (\S\ref{sect:highResResults}) we do not have a strong reason to expect different trends among fixed-order languages in a low-resource setup. 
We choose SOV because it is a commonly attested word order and is different from that of the target language, thereby requiring some non-trivial reorderings during translation.} and exclude syncretic case marking.
To disentagle the effect of word order from that of case marking on low-resource translation quality, we also experiment with a language variant combining fixed-order (SOV) and case marking.
Results are shown in Figure~\ref{fig:datasize} and discussed below.

\paragraph{Morphological Features}
%\subsection{Exponence and Flexivity}

% FROM WALS: Values of Map 21A. Exponence of Selected Inflectional Formatives
% exponence case+number 8/162 = 5%
% exponence all types  8+6+2/162 = 10%

The artificial case systems used so far included easily separable suffixes with a 1:1 mapping between grammatical categories and morphemes (e.g. \textit{.nsubj.sg, .dobj.pl}) reminiscent of agglutinative morphologies.
Many world languages, however, do not comply to this 1:1 mapping principle
% CAMERA-READY ??
% For instance, about 10\% of the 162 languages reviewed by \newcite{wals-21-exponence} displayed some form of \textit{exponence}, meaning that a single morpheme could convey multiple grammatical categories (e.g., tense, number). Half of this 10\% displayed exponence of number and case (including e.g., Russian and Finnish).
% Another morphological source of meaning-form complexity is \textit{flexivity}, whereby different suffixes can convey the same grammatical category (or combination thereof) and the choice is determined by the lexicon \cite{bickel-nichols-2007}. Typical examples of flexive languages include Latin and Russian.
but display flexivity (multiple categories conveyed by one morpheme) % \textit{many:1}) 
and/or exponence (the same category expressed by various, lexically determined, morphemes). % \textit{1:many}).
Well-studied examples of languages with case+number exponence include Russian and Finnish, while flexive languages include, again, Russian and Latin.
Motivated by previous findings on the impact of fine-grained morphological features on language modeling difficulty \cite{gerz-etal-2018-relation}, %Finally, for the unambiguous case marking system, 
we experiment with three types of suffixes (see examples in Table~\ref{tab:cases}): 
\begin{itemize}
    \item \textbf{overt}: number and case are denoted by easily separable suffixes (e.g. \textit{.nsubj.sg, .dobj.pl}) similar to agglutinative languages (1:1);
    \item \textbf{implicit}: the combination of number and case is expressed by unique suffixes without internal structure (e.g. \textit{kar} for \textit{.nsubj.sg}, \textit{ker} for \textit{.dobj.pl}) similar to fusional languages. This system displays exponence (many:1);
    \item \textbf{implicit with declensions}: like the previous, but with three different paradigms each arbitrarily assigned to a different subset of the lexicon. This system displays exponence \textit{and} flexivity (many:many).
    %This phenomenon, also common in fusional languages, is called flexivity and leads to a more ambiguous mapping between morphemes and grammatical functions \cite{gerz-etal-2018-relation}.
\end{itemize}
A complete overview of our morphological paradigms is provided in Appendix~\ref{app:suffixes}
All our languages have moderate inflectional synthesis and, in terms of fusion, are exclusively concatenative. Despite this, the effect on vocabulary size is substantial: 180\% increase by overt and implicit case marking, 250\% by implicit marking with declensions (in the full data setting).

% CAMERA-READY?
% \footnote{Clearly, many more feature combinations could be evaluated, however extensively studying the impact of fine-grained morphological features on NMT quality is beyond the scope of this paper.}

%\AB{mention dictionary size when adding cases}
%wc -l *n-dict*
%%  138085 en_randomna-aseed5-fr.train-dict
%  250933 en_randomna-d-f3seed5-fr.train-dict +2.5(none)
%  180737 en_randomna-d-fseed5-fr.train-dict
%  180766 en_randomna-dseed5-fr.train-dict =1.8(none)
%  101342 en_randomnoneseed5-fr.train-dict

%  72658 en_randomna-d-f3seed5-fr.train100k-dict
%  57802 en_randomna-d-fseed5-fr.train100k-dict
%  57810 en_randomna-dseed5-fr.train100k-dict
%  32046 en_randomnoneseed5-fr.train100k-dict

%  23045 en_randomna-d-f3seed5-fr.train10k-dict
%  20401 en_randomna-d-fseed5-fr.train10k-dict
%  20402 en_randomna-dseed5-fr.train10k-dict
%  12520 en_randomnoneseed5-fr.train10k-dict

\paragraph{Results}
Results are shown in the plots of Figure~\ref{fig:datasize} (detailed numerical scores are given in Appendix~\ref{app:figure2}). 
We find that reducing training size has, not surprisingly, a major effect on translation quality.
Among source language variants, fixed-order obtains the highest quality across all setups. %(tie in high-resource)
In terms of BLEU (\ref{fig:eparl-bleu}), the spread among variants increases somewhat with less data
%(with case marking never leading to better score than just random,no-case)
however differences are small.
A clearer picture emerges from RIBES (\ref{fig:eparl-ribes}), whereby less data clearly leads to more disparity. This is already visible in the 100k setup, with the fixed SOV language dominating the others.
Case marking, despite being necessary to disambiguate argument roles in the absence of semantic cues, does not improve translation quality and even degrades it in the low-resource setup. % leads to worse results on Europarl %(even in terms of RIBES) 
%
%Above all, data sparsity affects the ability of our models to understand and utilize source language structure as measured by the challenge set. 
Looking at the challenge set results (\ref{fig:control-ribes}) we see that the free-order case-marking languages are clearly disadvantaged: 
In the mid-resource setup, case marking improves substantially over the underspecified \textit{random,no-case} language but remains far behind fixed-order. % ($-4.4$ RIBES).
In low-resource, case marking notably hurts quality even in comparison with the underspecified language. 
% EXTRA:
% In terms of BLEU, overt case-marking languages score 3.6 against 9.0 by fixed SVO on the challenge set (not shown in the plot).
These results thus demonstrate that free-order case-marking languages require more data than their fixed-order counterparts to be accurately translated by state-of-the-art NMT.\footnote{In the light of this finding, it would be interesting to revisit the evaluation of \newcite{bugliarello-etal-2020-easier} in relation to varying data sizes.}
%The comparison of SOV, SOV+overt, random and random+overt
Our experiments also show that this greater learning difficulty is not only due to case marking (and subsequent data sparsity), but also to word order flexibility (compare \textit{sov+overt} to \textit{r+overt} in Figure \ref{fig:datasize}).

Regarding different morphology types, we do not observe a consistent trend in terms of overall translation quality (Europarl-test): 
in some cases, the richest morphology (with declensions) slightly outperforms the one without declensions ---a result that would deserve further exploration.
On the other hand, results on the challenge set, where most words are case-marked, show that morphological richness inversely correlates with translation quality when data is scarce. 
We postulate that our artificial morphologies may be too limited in scope (only 3-way case and number marking) to impact overall translation quality and leave the investigation of richer inflectional synthesis to future work.
%
%In future work, we plan to extend these experiments to other synthetic morphologies and particularly to richer inflectional synthesis, whose effect on vocabulary size is even stronger than that of the features considered here.

%\AB{refer to low res sizes in Conneau et al}
%wc -l -w
%   100000   2862248 src_train.100000.txt
%    10000    283842 src_train.10000.txt
%  1900419  54388812 src_train.txt
%   100000   3086357 tgt_train.100000.txt
%    10000    306558 tgt_train.10000.txt
%  1900419  58698002 tgt_train.txt

%\AB{- Gerz also studied Fusion and Inflectional Synthesis: the first has weak, the 2nd as moderate correlation with LM perplexity because it has direct effect on TTR.
%It would be interesting to study effect of infl. synthesis however here we only play with few grammatical functions, so that's out of scope}

%\subsection{Effect of Word Segmentation}
%\AB{one extra exp (with 100k) where we perfectly split suffixes from words prior to BPE. Could go with prev subsection}

\subsection{Effect of Target Language}

All results so far involved translation \textit{into} a fixed-order (SVO) language without case marking. To verify the generality of our findings, we repeat a subset of experiments with the same synthetic English variants, but using Czech or Dutch as target languages.
Czech has rich fusional morphology including case marking, and very flexible order.
Dutch has simple morphology (no case marking) and moderately flexible, syntactically determined order.\footnote{Dutch word order is very similar to German, with the position of S, V, and O depending on the type of clause.}
% from Are All Languages Equally Hard to Language-Model ?
% MCC:
% CS = 195
% NL = 26
% FR = 30
% EN = 6

\begin{figure}[ht]
%\centering
\subfigure[Europarl BLEU]{\hspace{-0.25cm}
    \includegraphics[width=.5\columnwidth, keepaspectratio]{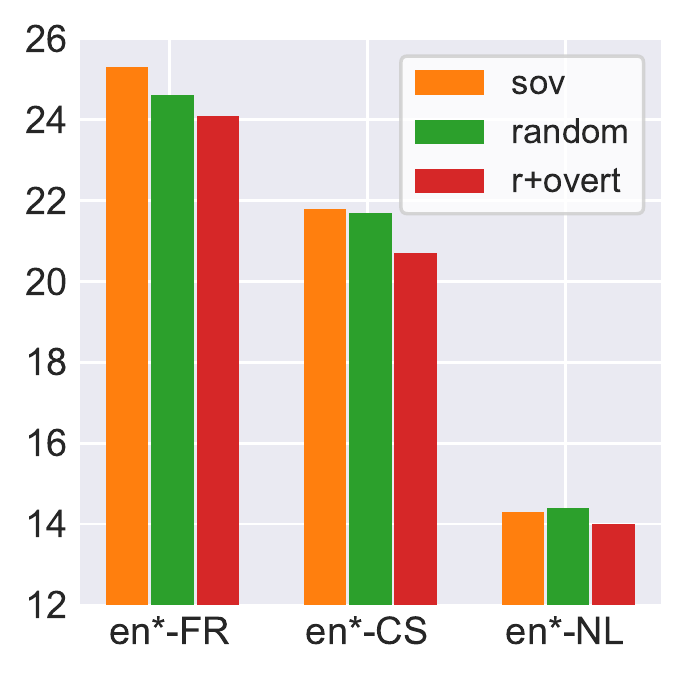}    
    \label{fig:tgt-eparl-bleu}
}\hspace{-0.24cm}
\subfigure[Europarl RIBES]{
    \includegraphics[width=.5\columnwidth, keepaspectratio]{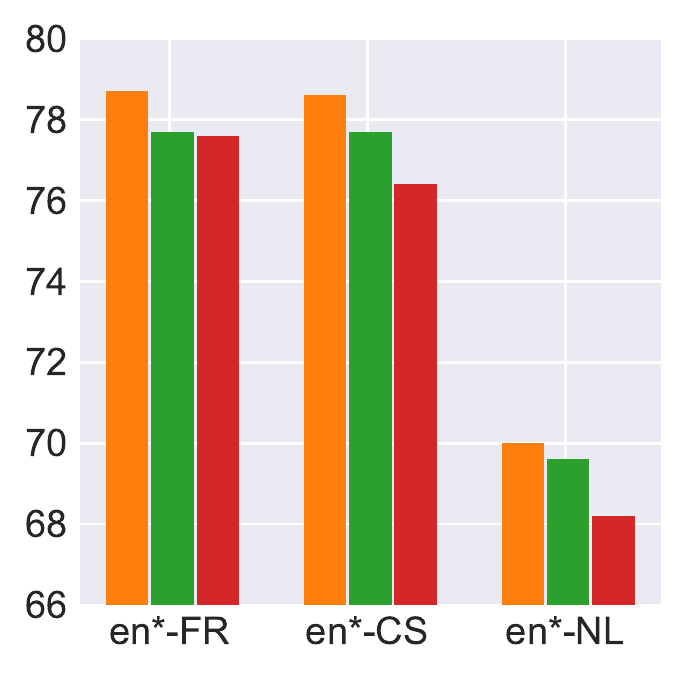} 
    \label{fig:tgt-eparl-ribes}
}
\caption{Transformer results for more target languages (100k training size). Scores averaged over 2 runs.}
\label{fig:target-langs}
\end{figure}

Figure~\ref{fig:target-langs} shows the results with 100k training sentences.
In terms of BLEU, differences are even smaller than in English-French.
In terms of RIBES, trends are similar across target languages, with the fixed SOV source language obtaining best results and the case-marked source language obtaining worst results. %
This suggests that the major findings of our study are not due to the specific choice of French as the target language.

\section{Related Work}

The effect of word order flexibility on NLP model performance has been mostly studied in the field of \textbf{syntactic parsing}, for instance using Average Dependency Length \cite{Gildea:2010,Futrell10336} or head-dependent order entropy \cite{futrell-15-quantifying,gulordava-merlo-2016-multi} as syntactic correlates of word order freedom.
Related work in \textbf{language modeling} has shown that certain languages are intrinsically more difficult to model than others \cite{cotterell-etal-2018,mielke-etal-2019} and has furthermore studied the impact of fine-grained morphology features %like fusion and exponence
\cite{gerz-etal-2018-relation} on LM perplexity.

Regarding the \textbf{word order biases of seq-to-seq models},
\newcite{Chaabouni2019} use miniature languages similar to those of Sect.~\ref{sect:expStephan} %for a different research question -- namely 
to study the evolution of LSTM-based agents in a simulated iterated learning setup. % display a trade-off between different ways to encode constituent roles similar to natural languages. %While such a trade-off is not observed, 
Their results in a standard ``individual learning'' setup show, like ours, that a free-order case-marking toy language can be learned just as well as a fixed-order one, confirming earlier results obtained by simple Elman networks trained for grammatical role classification \cite{lupyan2002case}. 
Transformer was not included in these studies.
\newcite{choshen-abend-2019-automatically} measure the ability of LSTM- and Transformer-based NMT to model a language pair where the same arbitrary (non syntactically motivated) permutation is applied to all source sentences. They find that Transformer is largely indifferent to the order of source words (provided this is fixed and consistent across training and test set) but nonetheless struggles to translate long dependencies actually occurring in natural data.
They do not directly study the effect of order flexibility.
%, but interestingly find that the same Transformer models struggle to translate long dependencies actually occurring in natural data.
%"but also shows that this does not entail that the model can overcome the difficulties of LDD in naturalistic data
%"phenomena are especially challenging due to the non-standard linguistic structure (e.g., syntactic and lexical structure), and the varying distances in which LDD manifest themselves. The

%Our 'Random, no case' language is clearly related, but arguably closer to the actual order flexibility

%- \cite{Du2017} \cite{Zhao2018} preordering for NMT

The idea of permuting dependency trees to generate synthetic languages was introduced independently by \newcite{gulordava-merlo-2016-multi} (discussed above) %with the aim of minimizing dependency length and word order flexibility, 
and by \newcite{wang-eisner-2016-galactic}, the latter with the aim of diversifying the set of treebanks currently available for language adaptation.

% From Galactic TB paper: Machine translation researchers have often tried to automatically preprocess parse trees of a source language to more closely resemble those of the tar- get language, using either hand-crafted or automati- cally extracted rules (Dorr et al., 2002; Collins et al., 2005, etc.; see review by Howlett and Dras, 2011)

% EXTRA STUFF, INTERESTING FOR LATER:
%{cite here or in Rel Work, subsect on word order evolution? \cite{Gell-Mann17290}}
%\AB{cite Dieuwke's JAIR about need for artificial data}
%\AB{check this: On the Computational Power of Transformers and Its Implications in Sequence Modeling Satwik Bhattamishra, Arkil Patel and Navin Goyal}
%\AB{as example of inconclusive results due to limited language sample cite: Processing effort is a poor predictor of cross-linguistic word order frequency. Gonering Morgan}

\section{Conclusions}
We have presented an in-depth analysis of how Neural Machine Translation difficulty is affected by word order flexibility and case marking in the source language.
Although these common language properties were previously shown to negatively affect parsing and agreement prediction accuracy, our main results show that state-of-the-art NMT models, especially Transformer-based ones, have little or no bias towards fixed-order languages.
Our simulated low-resource experiments, however, reveal a different picture, that is: free-order case-marking languages %are more affected by data scarcity and 
require more data to be translated as accurately as their fixed-order counterparts.  %by modern NMT systems.
Since parallel data (like labeled data in general) is scarce for most of the world languages \cite{guzman2019flores,joshi-etal-2020-state}, we believe this should be considered as a further obstacle to language equality in future NLP technologies. 

In future work, our analysis should be extended to target language variants using principled alternatives to BLEU \cite{bugliarello-etal-2020-easier}, and to
other typological features that are likely to affect MT performance, such as inflectional synthesis and degree of fusion \cite{gerz-etal-2018-relation}.
Finally, the synthetic languages and challenge set proposed in this paper could be used to evaluate syntax-aware NMT models \cite{eriguchi-etal-2016-tree,bisk-tran-2018-inducing,currey-heafield2019-tree}, which promise to better capture linguistic structure, especially in low-resource scenarios.

\section*{Acknowledgements}
Arianna Bisazza was partly funded by the Netherlands Organization for Scientific Research (NWO) under project number 639.021.646. We would like to thank the Center for Information Technology of the University of Groningen for providing access to the Peregrine HPC cluster, and the anonymous reviewers for their helpful comments.

%%%%%%%%%%%%%%%%%%%%%%%%%%%%%%%%%%%
\bibliography{anthology,tacl2018,thesis_stephan}
%\bibliography{tacl2018,thesis_stephan}
\bibliographystyle{acl_natbib}
%%%%%%%%%%%%%%%%%%%%%%%%%%%%%%%%%%%

\vspace{1mm}

\appendix
\section{Appendices}
\label{sec:appendix}

% CAMERA READY: Restore \subsections
\subsection{NMT Hyperparameters}
\label{app:hparams}

In the toy parallel grammar experiments (\S\ref{sect:expStephan}), batch size of 64 (sentences) and 1K max update steps are used for all models. We train BiLSTM with learning rate 1, %0.001
and Transformer with %starting 
learning rate of 2 together with 40 warm-up steps by using noam learning rate decay. Dropout ratio of 0.3 and 0.1 are used in BiLSTM and Transformer models respectively. In the synthetic English variants experiments (\S\ref{sect:expSynth}), we set a constant learning rate of 0.001 for BiLSTM. We also increased batch size to 128, number of warm-up steps to 80K and update steps to 2M for all models. Finally, for 100k and 10k datasize experiments, we decreased the warm-up steps to 4K. %For all experiments we used 32k BPE operations except 10k datasize experiment in which we decreased the number to 10k. 
During evaluation we chose the best performing model on validation set.

\iffalse
\begin{table}[h]
\centering \small
\begin{tabular}{ lllll}
& \bf TG & \bf 1.9M En & \bf 100K En & \bf 10K En \\
\midrule
Batch size & 64 & 128 & 128 & 128  \\
Warmup steps & 40 & 80k & 4k & 4k  \\
Train steps & 10k & 2m & 200k & 200k  \\
Bpe & 32k & 32k & 32k & 10k  \\
\end{tabular}
\caption{Hyperparameters used in experiments}
\label{tab:hyperparams}
\end{table}
\fi

% AB: STUFF FOR APPENDIX
%Because the 2-layer LSTM achieved 100\% accuracy using the default configuration of 100~000 training steps, we lower this number to 1~000 steps for each of our models.
%1~000 steps was chosen because all LSTM models were able to achieve 100\% accuracy within this amount of steps.
%As other settings are related to the number of training steps, we also reduce the number of validation steps from 10~000 to 100, the number of warm up steps from 4~000 to 40 and the number of checkpoint steps from 5~000 to 50.
%For the Transformer models, we reduce the batch size from 4~096 to 64, the batch type and normalization method from tokens to sentences and the optimization method from Adam \citep{Kingma2015} to stochastic gradient descent (SGD) to make these settings equal to the ones we use in our LSTM models.

\subsection{Challenge Set}
\label{app:challenge}
The English-French challenge set used in this paper, and available at \url{https://github.com/arianna-bis/freeorder-mt}, is generated by a small synchronous context-free grammar and contains 7,200 simple sentences consisting of a subject, a transitive verb, and an object (see Table~\ref{tab:challengeVocab}).
All sentences are in the present tense; half are affirmative, and half negative.
All nouns in the grammar can plausibly act as both subject and object of the verbs, so that an MT system must rely on sentence structure to get perfect translation accuracy.
The sentences are from a general domain, but we specifically choose nouns and verbs with little translation ambiguity that are well represented in the Europarl corpus: % (on both English and French side): 
most have thousands of occurrences, while the rarest word has about 80. 
Sentence example (English side): 
\textit{`The teacher does not respect the student.'} %/L'enseignant ne respecte pas l'étudiant.} 
and its reverse: \textit{`The student does not respect the teacher.'} %/L'étudiant ne respecte pas l'enseignant.}

%The following are example sentences taken from the set, in the original English version: 
%\AB{REMOVE THIS IF NEED SPACE!}
%\eenumsentence{\small
%\item The presidents thank the ministers. /\\ Les présidents remercient les ministres.
%\item The ministers thank the presidents. /\\ Les ministres remercient les présidents.
%}
%\eenumsentence{\small
%\item The teacher does not respect the student. /\\ 
%L'enseignant ne respecte pas l'étudiant.
%\item The student does not respect the teacher. /\\
%L'étudiant ne respecte pas l'enseignant.
%}

\begin{table}[h]
\centering \small
\begin{tabular}{ l|l}
\bf Nouns & \bf Verbs \\
\midrule
president / président     &  thank / remercier  \\
man / homme     &  support / soutenir \\
woman / femme     & represent / représenter \\
minister / ministre     & defend / défendre \\
candidate / candidat     & welcome / saluer \\
secretary / secrétaire     & invite / inviter \\
commissioner / commissaire     & attack / attaquer \\
child / enfant     & respect / respecter \\
teacher / enseignant     & replace / remplacer \\
student / étudiant     & exploit / exploiter \\
\end{tabular}
\caption{The English/French vocabulary used to generate the challenge set. Both singular and plural forms are used for each noun.}
\label{tab:challengeVocab}
\end{table}

\subsection{Morphological Paradigms}
\label{app:suffixes}

The complete list of morphological paradigms used in this work is shown in Table~\ref{tab:paradigms}.
The implicit language with exponence (many:1) uses only the suffixes of the 1$^{st}$ (default) declension.
The implicit language with exponence and flexivity (many:many) uses three declensions, assigned as follows:
First, the list of lemmas extracted from the training set is randomly split into three classes,\footnote{See \cite{williams-2020-declensions} for an interesting account of how declension classes are actually partly predictable from form and meaning.} with distribution 1$^{st}$:60\%, 2$^{nd}$:30\%, 3$^{rd}$:10\%.
Then, each core verb argument occurring in the corpus is marked with the suffix corresponding to its lemma's declension.

\begin{table}[h]
\centering \small 
\begin{tabular}{c  l @{\ \ \ \ \ \ \ } l @{\ \ }l l}
    & \multirow{2}{*}{\bf \ \ Overt} & \multicolumn{3}{c}{\bf Implicit} \\
    &   & {\bf 1$^{st}$}(default) & \bf 2$^{nd}$ & \bf 3$^{rd}$ \\
\hline
\multirow{6}{*}{\bf Unambiguous} & .nsubj.sg & kar & par & pa \\
    & .nsubj.pl & kon & pon & po \\
    & .dobj.sg & kin & it & kit \\
    & .dobj.pl & ker & et & ket \\
    & .iobj.sg & ken & kez & ke \\
    & .iobj.pl & kre & kr & re \\
\hline
\multirow{2}{*}{\bf Syncretic} & .arg.sg  % & kaz & kz & ka \\
& -- & -- & -- \\
    & .arg.pl  %& koz & kz & ko \\
    & -- & -- & -- \\
\midrule
\end{tabular}
\caption{The artificial morphological paradigms used in this work, extended from \cite{Ravfogel2019}. 1$^{st}$, 2$^{nd}$ and  3$^{rd}$ are the declensions in the flexive language.}
\label{tab:paradigms}
\end{table}

%.nsubj.sg .nsubj.sg .dobj.sg .dobj.sg .iobj.sg .iobj.sg .arg.sg .arg.sg

\subsection{Effect of Data Size and Morphological Features: Detailed Results}
\label{app:figure2}

Table~\ref{tab:figure2} shows the detailed numerical results corresponding to the plots of Figure~\ref{fig:datasize} in the main text.

% \usepackage{booktabs}
\begin{table}[h]
\centering\small
\begin{tabular}{@{}llll@{}}
\toprule
\textit{Eparl-BLEU}    & 1.9M & 100k & 10k  \\ 
\hdashline
original      & 38.3      & 26.9 & 11.0 \\
SOV           & 37.9      & 25.3 & 8.8  \\
SOV+overt           & 37.4      & 24.6 & 8.4  \\
random        & 37.5      & 24.6 & 8.5  \\
random+overt        & 37.3      & 24.1 & 7.8  \\
random+implicit      & 37.3      & 24.3 & 7.1  \\
random+declens     & 37.4      & 23.1 & 7.7  \\
\midrule
\textit{Eparl-RIBES}   & 1.9M & 100k & 10k  \\
\hdashline
original      & 84.9      & 80.1 & 67.5 \\
SOV           & 84.5      & 78.7 & 64.1 \\
SOV+overt           & 84.5      & 78.4 & 63.1  \\
random        & 84.2      & 77.7 & 61.7 \\
random+overt        & 84.4      & 77.6 & 61.6 \\
random+implicit      & 84.3      & 77.4 & 59.8 \\
random+declens     & 84.3      & 77.1 & 61.3 \\
\midrule
\textit{Challenge-RIBES} & 1.9M & 100k & 10k  \\
\hdashline
original      & 97.7      & 92.2 & 74.2 \\
SOV           & 97.2      & 89.8 & 69.5 \\
SOV+overt           & 97.7      & 86.5 &  64.9  \\
random        & 74.1      & 72.9 & 63.1 \\
random+overt        & 98.1      & 84.5 & 57.4 \\
random+implicit      & 97.5      & 85.4 & 54.8 \\
random+declens     & 97.6      & 84.4 & 53.1 \\ \bottomrule
\end{tabular}
\caption{Detailed results corresponding to the plots of Figure ~\ref{fig:datasize}: EN*-FR Transformer NMT quality versus training data size (1.9M, 100K, or 10K sentence pairs). 
Source language variants: Fixed-order (SOV) and free-order (random) with different case systems (+overt/implicit/declens). Scores averaged over three training runs.}
\label{tab:figure2}
\end{table}
    
%\subsection{}
%\label{app:challengeBleu}
%On the challenge set we observe larger differences but \AB{based on a manual check TODO} we attribute that to noise ...The RIBES metric is more robust to lexical mismatches and better reflect the true performance of our models.

\end{document}